%% file: neurips_2026.tex
\documentclass{article}
\PassOptionsToPackage{numbers, compress}{natbib}

\usepackage{amsmath}
\usepackage{pifont}
\usepackage{caption}
\usepackage[most]{tcolorbox}

\tcbset{
  promptbox/.style={
    enhanced,
    breakable,
    colback=gray!3,
    colframe=gray!35,
    boxrule=0.4pt,
    arc=1mm,
    left=5pt,
    right=5pt,
    top=5pt,
    bottom=5pt,
    fonttitle=\bfseries,
    coltitle=black,
    colbacktitle=gray!12
  }
}

\usepackage{enumitem}
\usepackage{wrapfig}
\newcommand{\cmark}{\ding{51}}
\newcommand{\xmark}{\ding{55}}
\usepackage[preprint]{neurips_2026}

\usepackage[utf8]{inputenc} 
\usepackage[T1]{fontenc}    
\usepackage{hyperref}       
\usepackage{url}            
\usepackage{booktabs}       
\usepackage{amsfonts}       
\usepackage{nicefrac}       
\usepackage{microtype}      
\usepackage{xcolor}         
\usepackage{enumitem}
\usepackage{graphicx}
\usepackage[table]{xcolor}
\usepackage{booktabs}
\usepackage{xspace}

\newcommand{\methodname}{DriveWAM\xspace}

\title{DriveWAM: Video Generative Priors Enable Scalable World-Action Modeling for Autonomous Driving}

\author{
  \parbox{\textwidth}{\centering
    \vspace*{0.3cm}
    Chen Shi$^{1*}$ \quad Jinrui Xu$^{1*}$ \quad Shaoshuai Shi$^2$ \quad
    Kehua Sheng$^2$ \quad Bo Zhang$^2$ \quad Li Jiang$^{1\dagger}$ \\[1ex]
    {\normalfont \small $^1$The Chinese University of Hong Kong, Shenzhen \quad $^2$Voyager Research, Didi Chuxing} \\[2ex]
    {\normalfont \textbf{Project Page:} \href{https://chenshi3.github.io/drivewam.github.io/}{\small\texttt{https://chenshi3.github.io/drivewam.github.io/}}}
  }
}

\begin{document}

\maketitle
\begingroup
\renewcommand{\thefootnote}{}
\footnotetext{${*}$:Equal Contribution. Work done during an internship at Voyager Research, Didi Chuxing. $\dagger$: Corresponding author.}
\endgroup

\begin{abstract}
Pretrained foundation models have become an important basis for end-to-end autonomous driving. In contrast to vision-language models pretrained primarily on static image-text pairs, video generative models capture temporal dynamics and motion priors that are naturally suited for driving. We present \methodname, a driving world-action model that adapts a pretrained video diffusion transformer into an autoregressive video-action policy. \methodname organizes video and action streams into a unified temporal token sequence and trains them under a joint flow-matching objective, preserving the pretrained video-generation architecture while adapting its large-scale video priors to action generation. To incorporate high-level scene understanding, we introduce scene-evolving driving guidance, where a frozen VLM produces chunk-specific semantic intent to guide video-action generation. To keep long-horizon rollout bounded, we further introduce selective KV memory, which maintains bounded modality-aware video and action memory pools through relevance-redundancy cache selection at inference time. Experiments on NAVSIM and the PhysicalAI-Autonomous-Vehicles benchmark show that \methodname achieves strong planning performance, and a data-scaling study from 4k to 100k driving clips further confirms the scaling potential of world-action modeling for end-to-end autonomous driving.
\end{abstract}
\input{section/intro_v3}

\input{section/related_work}

\input{section/method_v2}
\input{section/experimets}

{
    \small
    \bibliographystyle{unsrtnat}
    \bibliography{neurips_2026}
}

\newpage
\appendix
\input{section/appendix}




\end{document}

%% file: section/intro_v3.tex
\section{Introduction}

Recent end-to-end autonomous driving systems increasingly leverage pretrained foundation models as policy backbones. 
A major line of work builds on vision-language-action (VLA) models~\cite{black2024pi_0,intelligence2025pi_,li2026drivevlaw,li2025recogdrive,renz2025simlingo,yang2025drivemoe,gao2025langcoop}, transferring the semantic knowledge and instruction-following ability of large-scale VLMs~\cite{wang2024emu3,Qwen3-VL,liu2023visual,beyer2024paligemma,alayrac2022flamingo,wang2024qwen2} to action generation. 
Such VLA-based policies are well suited to high-level scene understanding and semantic reasoning, but driving decisions also require temporally dense visual cues such as spatial layout, motion continuity, and how the scene may evolve in the near future. Since VLM backbones are pretrained primarily on image-text data rather than video dynamics, VLM-centric policies must acquire these temporal priors largely from downstream driving data.

Video generative models offer a complementary foundation. 
They are pretrained on large-scale videos to model object persistence, motion patterns, and scene evolution, making them naturally suited for dynamic decision problems. 
Recent VLA-based driving methods~\cite{li2026drivevlaw,zhou2026drivedreamer,zeng2026futuresightdrive} have begun to incorporate future image or video generation to improve spatio-temporal awareness, but visual generation is often used as an auxiliary signal or a modular component on top of a VLM-centric policy. 
In parallel, world-action (WA) models in robotics~\cite{kim2026cosmos,ye2026world,yuan2026fastwam,li2026causal,ye2026gigaworld,hu2025video} show that pretrained video foundation models can be adapted more directly for action prediction and planning. 


Adapting this paradigm to autonomous driving, however, remains non-trivial. 
First, a video foundation model is pretrained for visual generation rather than ego-action control, so turning it into an autoregressive video-action policy requires preserving its future-generation prior while coupling it to continuous action prediction. Second, video foundation models capture near-future dynamics but lack high-level semantic planning, whereas the appropriate driving decision depends on route intent, right-of-way, and decision-relevant traffic participants. 
Third, deploying such autoregressive policies over long horizons requires persistent historical context, but full KV caching grows with horizon length and sliding-window caching may discard old yet critical evidence. 
Existing driving-oriented world-action methods~\cite{gui2026bridging,vavam2025,zhang2025epona} often rely on separate planners, discrete video tokenizers, or customized generation architectures, leaving open how to directly adapt a modern video foundation model into a semantically guided and scalable end-to-end driving policy.

In this paper, we present \methodname, a driving world-action model that adapts a pretrained video foundation model into an end-to-end autonomous driving policy. 
\methodname uses a flow-matching video diffusion transformer as the policy core and formulates driving as autoregressive video-action generation. 
Given observed video-action history and ego state, the model first generates future video latents and then decodes ego actions conditioned on the generated future latent, realizing inverse-dynamics action generation. Both video and action streams share the same transformer and are trained under a joint flow-matching objective~\cite{lipman2023flow}, preserving the pretrained spatio-temporal generative prior while learning to convert imagined future world evolution into executable ego motion.

To supply the missing high-level driving semantics, \methodname introduces scene-evolving driving guidance. A frozen VLM uses only causally available context, including the latest observation, recent ego motion, and route command, and produces chunk-specific guidance for the next prediction horizon. 
This guidance is injected through temporally localized cross-attention, ensuring that each future video-action chunk receives its own semantic intent while preserving the causal structure of full-clip autoregressive training. 
Thus, the VLM acts as a semantic guide, while the video foundation model remains responsible for dense temporal prediction.

For long-horizon rollout, \methodname further introduces selective KV memory. 
Instead of storing all historical tokens or evicting tokens by age, \methodname maintains separate bounded memory pools for video and action KVs. 
Each pool is updated by a relevance-redundancy selection rule inspired by efficient video-generation caching~\cite{ma2026flow}: prediction-relevant tokens are retained, while redundant patterns are filtered out. This training-free memory provides a compact video-action history for autoregressive inference without changing the training objective.

We evaluate \methodname on NAVSIM~\cite{dauner2024navsim} and the large-scale PhysicalAI-Autonomous-Vehicles benchmark~\cite{wang2025alpamayo}. 
\methodname achieves strong planning performance with an autoregressive world-action architecture. Beyond benchmark comparison, we conduct a data-scaling study over $4$k, $20$k, and $100$k driving clips, where \methodname improves consistently as training data increases. 
These results suggest that semantically guided world-action modeling provides a scalable foundation for end-to-end autonomous driving. Our contributions are summarized as follows:

\begin{itemize}[leftmargin=2em]
    \item We propose \methodname, a driving world-action model that adapts a pretrained video diffusion transformer into an autoregressive video-action policy under a joint flow-matching objective.

    \item We introduce scene-evolving driving guidance to supply high-level driving semantics, where a frozen VLM provides causally available chunk-specific intent that guides video-action generation through temporally localized cross-attention.

    \item We propose selective KV memory for bounded long-horizon rollout, maintaining modality-aware video and action memory pools through relevance-redundancy cache selection at inference time.

    \item Experiments on NAVSIM and PhysicalAI-Autonomous-Vehicles, together with a scaling study from $4$k to $100$k clips, demonstrate the effectiveness and scalability of \methodname.
\end{itemize}

%% file: section/related_work.tex
\section{Related Work}

\subsection{Vision-Language-Action Models in Autonomous Driving}
Recent autonomous driving methods increasingly leverage the general knowledge and semantic reasoning capabilities of large vision-language models. Early efforts use LLMs or VLMs mainly as high-level reasoning modules~\citep{xu2024drivegpt4, DriveVLM,jiang2024senna,sima2024drivelm,wang2023drivemlm,mao2023language,wang2025omnidrive,hwang2024emma}, producing scene descriptions, maneuver suggestions, command tokens, or coarse trajectories that are further consumed by downstream planners. More recent VLA methods~\cite{yang2025drivemoe,zhou2026autovla,li2025recogdrive} move toward end-to-end action prediction by coupling VLM backbones with trajectory decoders or planning heads. DriveMoE~\citep{yang2025drivemoe} introduces an MoE-based policy head on top of a VLM to route different driving situations to specialized action experts. AutoVLA~\citep{zhou2026autovla} discretizes continuous trajectories into action primitives and casts driving policy learning as autoregressive token prediction. ReCogDrive~\citep{li2025recogdrive} combines VLM-based reasoning with a diffusion trajectory planner and further aligns the policy through imitation learning and reinforcement learning.

Building on this line, a parallel set of works incorporates visual world modeling into the VLA pipeline. FSDrive~\citep{zeng2026futuresightdrive} introduces future visual prediction as a visual reasoning process, while DriveVLA-W0~\citep{li2026drivevlaw} and DriveDreamer-Policy~\citep{zhou2026drivedreamer} augment VLM-based policies with generative world-model components~\cite{zheng2022movq,peebles2023scalable,wan2025wan,yang2025cogvideox}. Although these designs improve the spatio-temporal awareness of VLA-based driving, their policy core remains VLM-centric, with visual generation serving as an auxiliary branch rather than the policy backbone. In contrast, DriveWAM inherits a pretrained video generative model as the policy core to jointly model future world evolution and ego actions, while leveraging VLM reasoning as complementary scene-evolving guidance for high-level semantic intent.

\subsection{World-Action Models}
The world-action paradigm reuses pretrained video generative models as the foundation for policy learning. Recent works in robotic manipulation~\citep{kim2026cosmos,ye2026world,yuan2026fastwam,li2026causal,hu2025video} have shown that large-scale video pretraining can transfer favorably to action generation, motivating its adoption in autonomous driving. WorldDrive~\cite{gui2026bridging} transfers representations learned by a trajectory-aware driving world model to a downstream planner, bridging scene generation and planning but keeping planning as a separate module. VaViM/VaVAM~\cite{vavam2025} formulates autonomous driving as autoregressive video modeling with discrete VQ-VAE tokens~\citep{van2017neural} through a GPT-style transformer~\cite{radford2019language}, and extends the model with an action expert for trajectory prediction. Epona~\citep{zhang2025epona} couples a spatiotemporal transformer with twin diffusion transformers for separate next-frame generation and ego-trajectory prediction. While these designs establish important baselines, they do not directly adopt a modern video foundation model as a unified video-action policy backbone, and thus cannot fully inherit the latest pretrained video priors. DriveWAM instead builds directly on a pretrained video diffusion transformer and adapts both video and action streams under a unified flow-matching objective.

Moreover, existing driving-oriented world-action methods mostly rely on simple navigation commands as high-level guidance, leaving rich scene-level semantic reasoning largely unexplored. DriveWAM addresses this by injecting chunk-specific VLM guidance through temporally localized cross-attention. Efficient memory is another requirement for autoregressive video-action policies during long-horizon rollout, but prior models either use a limited observation window~\cite{kim2026cosmos,yuan2026fastwam} or maintain a standard KV cache~\cite{ye2026world,li2026causal} whose cost grows with the sequence length. Recent works on efficient autoregressive video generation explore sliding-window attention~\cite{huang2026self,chen2024diffusion,yin2025slow}, sparse attention~\cite{xi2025sparse,yang2026sparse}, and cache compression~\cite{ma2026flow,kahatapitiya2025adaptive}. DriveWAM adapts the relevance-redundancy criterion of FlowCache~\cite{ma2026flow} to maintain bounded video and action memory pools for long-horizon driving.

%% file: section/method_v2.tex
\section{Method}
\begin{figure}
  \centering
  \includegraphics[width=0.999\linewidth]{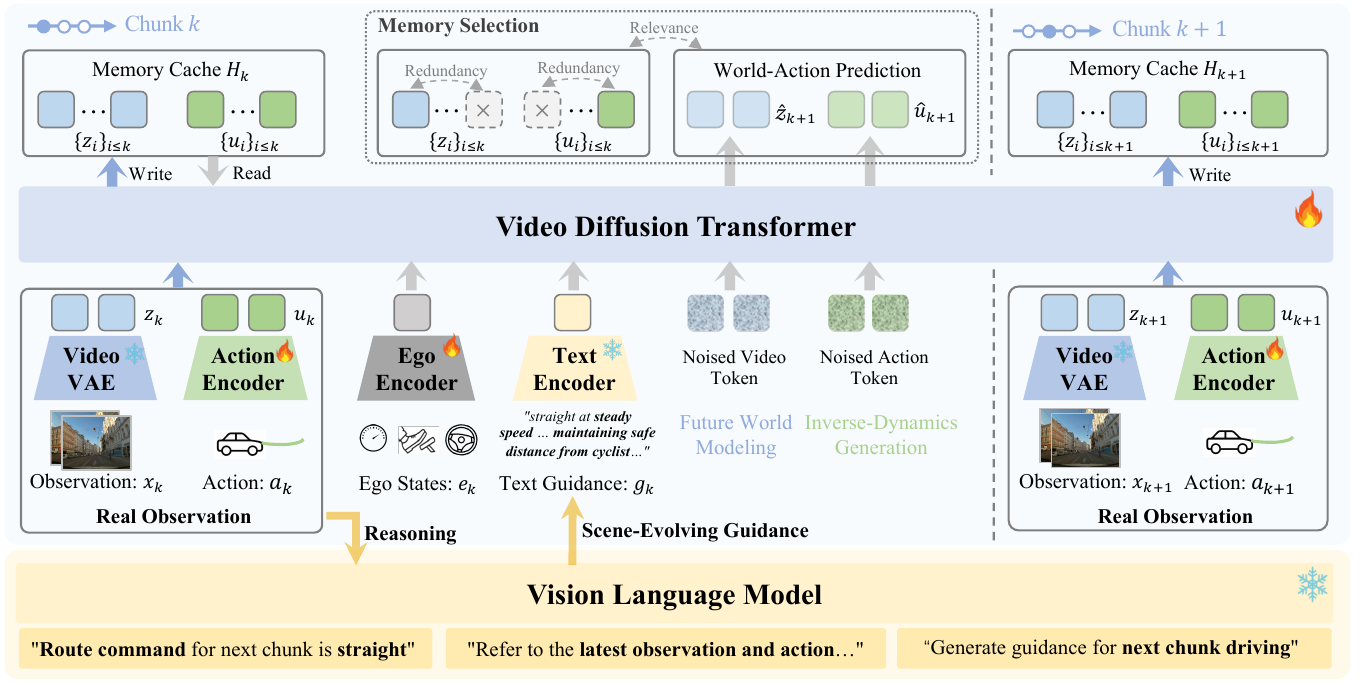}
  \caption{Overview of DriveWAM, which adapts a pretrained video generation backbone into a unified video-action policy. Building on this backbone, DriveWAM uses a frozen VLM to provide chunk-specific scene-evolving guidance for high-level scene reasoning and introduces selective KV memory to preserve compact prediction-relevant history for long-horizon rollout.}
  \label{fig:framework}
  \vspace{-8pt}
\end{figure}

We propose \methodname, a semantically guided world-action model that adapts a pretrained video foundation model into a unified backbone for future world evolution and ego-action generation in autonomous driving, complemented by guidance from a frozen VLM for scene-evolving driving semantics. Specifically, as shown in Figure~\ref{fig:framework}, we first formulate driving as autoregressive video-action generation, where a pretrained video diffusion transformer predicts future video latents and ego actions under a joint flow-matching objective (Sec.~\ref{sec:ava_formulation}). We then introduce scene-evolving guidance, using a frozen VLM to provide causally available chunk-level intent that steers the video-action generation process (Sec.~\ref{sec:vlm_guidance}). Finally, we present selective KV memory, which retains prediction-relevant and non-redundant video-action history for bounded long-horizon rollout (Sec.~\ref{sec:compressed_kv}).

\subsection{Autoregressive Video-Action Generation}\label{sec:ava_formulation}

A driving clip contains synchronized streams of camera images, ego actions, and ego states. We divide the clip into $K$ consecutive chunks and then consider the driving task as the next-chunk generation. At decision step $k$, the model has observed the clip up to chunk $k$ and predicts the future video-action chunk $(x_{k+1}, a_{k+1})$, where $x_{k+1}$ is the next video segment and $a_{k+1}$ is the corresponding ego action. The causally available conditions include the historical context $H_k$ (video and action tokens of all observed chunks up to $k$), the ego state $e_k$ at the end frame of chunk $k$ (\emph{e.g.}, velocity, acceleration, and curvature), and a textual guidance $g_k$ for the predicted chunk.

\paragraph{Tokenization.}
To jointly model video-action generation, we organize video and action chunks into a unified temporal token sequence while preserving their temporal order.
Each observed video chunk is encoded by the pretrained VAE~\cite{wan2025wan}, and each ego-action chunk, represented as normalized ego-frame translation and yaw increments, is embedded by an MLP action encoder $E_a$, as follows:
\begin{equation}
    z_k = \mathrm{VAE}(x_k), \qquad
    u_k = E_a(a_k), \qquad
    H_k = \{(z_i,u_i)\}_{i\leq k}.
\end{equation}
Here, $z_k\in\mathbb{R}^{N_x\times d_z}$ and
$u_k\in\mathbb{R}^{N_a\times d}$ denote encoded video and action tokens, respectively. $N_x$ and $N_a$ are the numbers of tokens per chunk, $d_z$ is the VAE latent channel dimension, and $d$ is the transformer hidden dimension. In practice, the VAE latents $z_k$ are also mapped to dimension $d$ by the latent input embedding layer inherited from the pretrained video diffusion transformer, yielding a unified representation for video-action generation.

\paragraph{World-action flow.} \methodname adopts the autoregressive video-action generation scheme, which factors the driving task into future world modeling and inverse-dynamics action generation. Specifically, \methodname utilizes a pretrained flow-matching video diffusion transformer $T_\omega$~\cite{wan2025wan} for predicting the next video chunk and action chunk. During training, we sample a flow timestep $\tau\in[0,1]$ along the rectified-flow path~\cite{lipman2023flow,liu2023flow}, where $\tau=1$ is the Gaussian-noise endpoint and $\tau=0$ represents clean data. For the next video chunk, the clean latent $z_{k+1}$ is noised along the standard rectified-flow path, producing a query $z_{k+1,\tau}$ and target velocity $v^z_{k+1,\tau}$. The video branch predicts this velocity under the current
driving context:
\begin{equation}
    \hat v^z_{k+1,\tau}
    = T_\omega(z_{k+1,\tau}; H_k, e_k, g_k, \tau).
\label{eq:video_flow}
\end{equation}
Here, $e_k$ is embedded by a lightweight MLP and injected through a separate ego-state cross-attention branch.  Notably, this conditioning repurposes the video model as a policy prior, with the backbone retaining its native future-visual-prediction objective while the predicted future is shaped by driving history, ego state, and semantic intent.

Actions are generated by an inverse-dynamics flow on the same diffusion transformer. We perturb the next action chunk directly in the normalized action space and embed it with the MLP action encoder $E_a$ to obtain $u_{k+1,\tau}$. 
Conditioned on the future world latent and the current driving context, the shared transformer predicts the action velocity as:
\begin{equation}
    \hat v^a_{k+1,\tau}
    =
    D_a\!\left(
    T_\omega(u_{k+1,\tau}; \tilde z_{k+1}, H_k, e_k, g_k, \tau)
    \right),
\label{eq:action_flow}
\end{equation}
where $D_a$ is an MLP action decoder. 
The conditioning latent $\tilde z_{k+1}$ is the clean future video latent $z_{k+1}$ during teacher-forced training and the generated latent $\hat z_{k+1}$ during inference. This design grounds action generation in the predicted world evolution, so the action decoder behaves as an inverse-dynamics readout of the predicted future rather than an independent trajectory head. 
We use noisy-history augmentation~\cite{li2026causal} to reduce this train-test mismatch.

\paragraph{Training objective.}
We train the video and action branches with a joint flow-matching objective:
\begin{equation}
    \mathcal{L}
    =
    \mathbb{E}_{k,\tau}
    \left[
    \left\|\hat v^z_{k+1,\tau}-v^z_{k+1,\tau}\right\|_2^2
    +
    \beta_a
    \left\|\hat v^a_{k+1,\tau}-v^a_{k+1,\tau}\right\|_2^2
    \right],
    \label{eq:joint_loss}
\end{equation}
where $\beta_a$ controls the balance between future world modeling and action generation. The video term preserves the pretrained spatio-temporal generative prior during policy adaptation, while the action term teaches the shared backbone to decode this prior into executable ego motion.

\paragraph{Full-clip training and autoregressive rollout.}
During training, we process all chunks of a clip in a single forward pass for efficiency. The video-action tokens are arranged in temporal order and denoised in parallel under a causal teacher-forcing mask (Figure~\ref{fig:attn_mask}), which realizes the conditional dependencies in Eqs.~\ref{eq:video_flow} and~\ref{eq:action_flow} while preserving the causal pattern used during inference~\cite{chen2024diffusion,li2026causal,huang2026self}. At inference, \methodname rolls out one chunk at a time.
Given history $H_k$, the model first samples the future video latent $\hat z_{k+1}$ and then samples the action chunk $\hat a_{k+1}$ conditioned on this generated future. When the next real observation becomes available, it is encoded and appended to the history to form $H_{k+1}$, keeping long-horizon rollout grounded in observed driving context.

\subsection{Scene-Evolving Driving Guidance}
\label{sec:vlm_guidance}
The video foundation model provides dense dynamic priors for near-future scene evolution, but it lacks semantic planning ability. In driving, the appropriate future is determined not only by short-term dynamics but also by route intent, traffic participants, and other decision-level semantics. For example, at an intersection, multiple future evolutions may be visually plausible from the current observation, while the desired one depends on the high-level driving intent. However, existing world-action methods typically use a single clip-level text condition, applying the same semantic guidance to every chunk. \methodname instead introduces a frozen VLM as a scene-evolving semantic guide. At each decision step $k$, the VLM produces fresh guidance $g_k$ from the latest causally available context, so each future video-action chunk is conditioned on its own up-to-date semantic intent while the video model remains the policy backbone for dense temporal prediction.

\paragraph{Causal guidance generation.}
At each decision step $k$, the frozen Qwen3-VL-8B~\cite{Qwen3-VL} receives only causally available information: the latest observation $x_k$, a recent ego trajectory $a_k$, and the route command $c_k$ for the upcoming horizon. It produces a concise guidance text as follows:
\begin{equation}
    g_k = \Phi_{\mathrm{VLM}}(x_k, a_k, c_k),
    \label{eq:vlm_guidance}
\end{equation}
which summarizes the current road context and provides ego behavior guidance for the upcoming horizon, such as proceeding, yielding, stopping, or merging. Since no observation from the target chunk is used, $g_k$ provides semantic intent for predicting $(x_{k+1},a_{k+1})$ without leaking future information. 
During training, guidance texts are precomputed and cached; during inference, the VLM is queried once per decision step and reused across all denoising steps, keeping the semantic condition aligned with the current prediction horizon.

\begin{figure}[t]
  \centering
  \begin{minipage}[t]{0.46\linewidth}
    \centering
    \includegraphics[width=\linewidth]{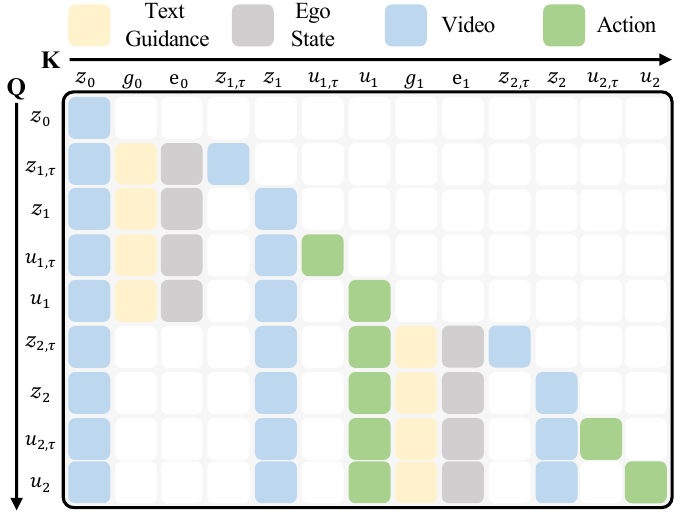}
    \captionof{figure}{Attention mask used during DriveWAM training. Colored entries indicate allowed attention; blank entries are masked.}
    \label{fig:attn_mask}
  \end{minipage}
  \hfill
  \begin{minipage}[t]{0.51\linewidth}
    \centering
    \centering
    \includegraphics[width=\linewidth]{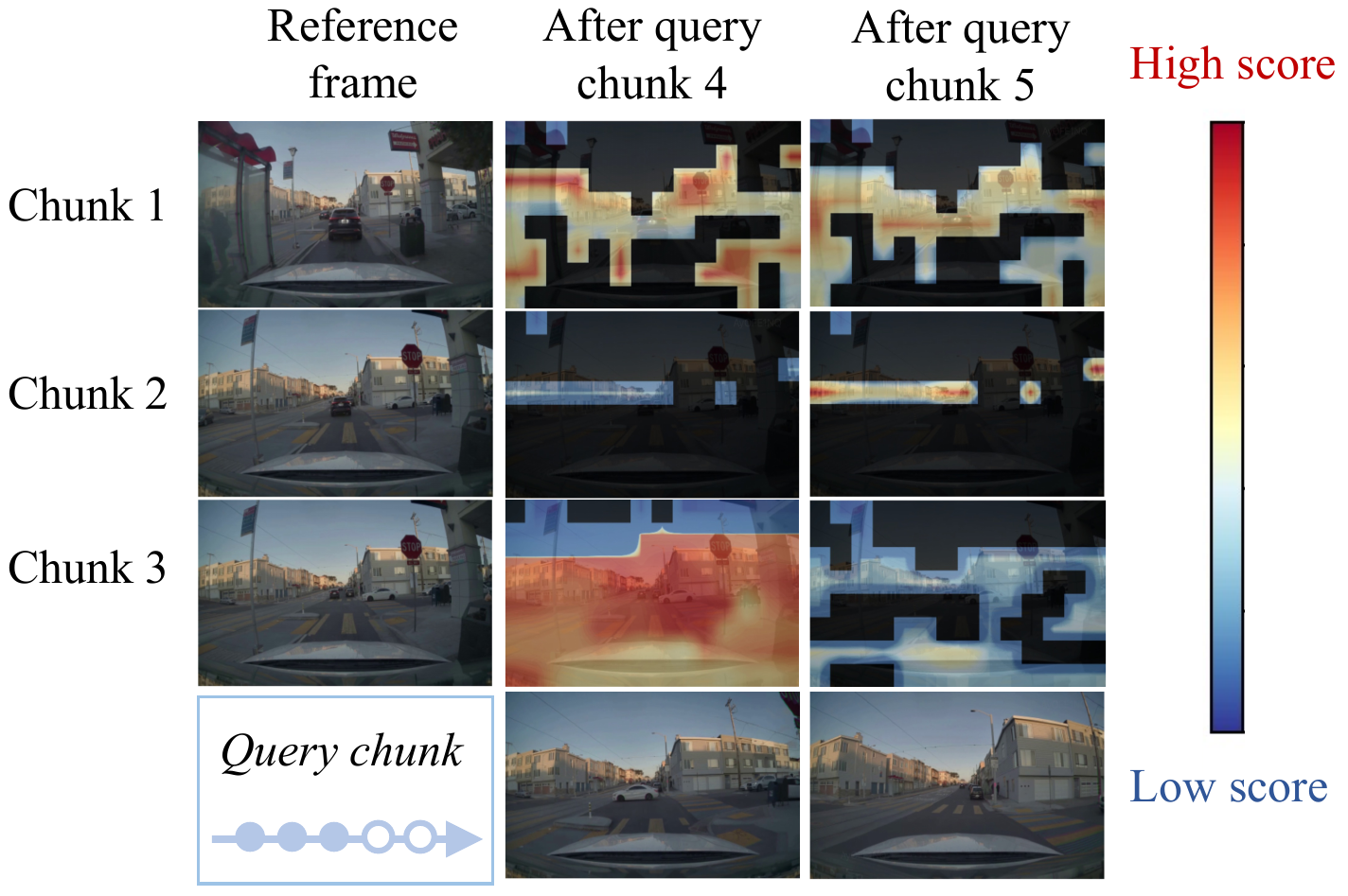}
    \captionof{figure}{Video-token retention under selective KV memory. Columns 2 and 3 visualize the tokens retained after query chunks 4 and 5.}
    \label{fig:kv_cache}
  \end{minipage}
  \vspace{-10pt}
\end{figure}

\paragraph{Temporally localized guidance injection.} Scene-evolving guidance introduces a separate text condition $g_k$ at each decision step. Without an additional constraint, tokens of chunk $k+1$ could attend to guidance from other chunks, including future guidance from later decision steps, breaking causal consistency. We therefore apply an additional block-diagonal text mask, which allows video-action tokens of target chunk $k+1$ to attend only to the guidance tokens of $g_k$. This keeps semantic conditioning temporally localized and prevents cross-chunk leakage. The resulting attention pattern is illustrated in Figure~\ref{fig:attn_mask}.

\subsection{Selective KV Memory for Long-Horizon Rollout}
\label{sec:compressed_kv}
Autoregressive world-action rollout conditions on the historical context $H_k$ defined in Sec.~\ref{sec:ava_formulation}, where $H_k$ denotes the causal video-action history up to step $k$. During inference, this abstract history is implemented as layer-wise KV caches that store the keys and values produced by previous video and action chunks, so the model can attend to past context without recomputing all historical tokens. A full-window cache preserves complete history but grows linearly with rollout length, while a sliding-window cache bounds the cost by evicting the oldest tokens under FIFO rules~\cite{kim2024fifodiffusion,zheng2026xworld}. However, age-based eviction is suboptimal for driving tasks: older tokens may remain decision-relevant, such as motion trend of a nearby vehicle or a briefly occluded pedestrian, whereas newer tokens may correspond to repeated static background. To keep long-horizon inference bounded without discarding useful context, \methodname adopts an inference-time, training-free selective KV memory inspired by FlowCache~\cite{ma2026flow}, retaining a compact, prediction-relevant approximation of $H_k$ during rollout.


\paragraph{Modality-aware memory pools.}
Video and action histories have different token densities and functional roles. 
Video tokens are numerous and encode scene context, while action tokens are compact and encode ego-motion history. 
A single global cache would therefore be dominated by visual tokens and may under-preserve motion context. \methodname decomposes $H_k$ into two bounded modality pools $H^v_k$ and $H^a_k$, with $|H^v_k|\leq B^v$ and $|H^a_k|\leq B^a$, where $B^v$ and $B^a$ are the video and action memory budgets. This modality-aware design keeps both scene evidence and ego-motion history available during long-horizon rollout.

\paragraph{Relevance-redundancy retention.}
When a memory pool exceeds its budget, \methodname ranks cached tokens by both current relevance and memory complementarity. 
For modality $m\in\{v,a\}$, let $Q_k^m$ denote the current query tokens of modality $m$, and let $\mathbf{k}^m_j$ be the cached key of token $j$ in $H^m_k$. We measure relevance {$\rho^m_j$} by the average attention mass assigned to token $j$ from current queries, and redundancy {$\eta^m_j$} by its average similarity to other cached keys:
\begin{equation}
    \rho^m_j =
    \frac{1}{|Q_k^m|}
    \sum_{\mathbf{q}\in Q_k^m}
    \left[
    \mathrm{softmax}_{\ell\in {H}_k^m}
    \left(
    \frac{\mathbf{q}^\top \mathbf{k}^m_\ell}{\sqrt{d}}
    \right)\right]_j,
    \qquad
    \eta^m_j =
    \mathrm{mean}_{\ell\neq j}\cos(\mathbf{k}^m_j,\mathbf{k}^m_\ell),
    \label{eq:memory_terms}
\end{equation}
where $d$ is the transformer hidden dimension. The final retention score is:
\begin{equation}
    s^m_j = \lambda \rho^m_j - (1-\lambda)\eta^m_j,
    \label{eq:retention}
\end{equation}
where $\lambda\in[0,1]$ balances relevance and redundancy, and tokens with low scores are evicted. As shown in Figure~\ref{fig:kv_cache}, this criterion has a natural driving-oriented interpretation: repeated road surfaces, sky, buildings, and other static background regions tend to be filtered out, while prediction-relevant cues such as moving vehicles and lane geometry are more likely to be retained.

\paragraph{Inference procedure.}
Selective KV memory is applied only at inference time and does not change the training objective or model parameters. 
During rollout, each transformer layer attends to the current chunk together with the bounded video and action memory pools. 
After chunk $k+1$ is processed, the existing memory is ranked by the retention score, and the lowest-scored historical tokens are evicted to make room for the newly generated KVs $\Delta H^m_{k+1}$.  The modality pool is then updated as:

\begin{equation}
    H^m_{k+1}    
    \leftarrow
    \mathrm{Top}_{B^m-|\Delta H^m_{k+1}|}
    \!\left(H^m_k\right)
    \cup
    \Delta H^m_{k+1},
    \qquad m\in\{v,a\}.
    \label{eq:memory_update}
\end{equation}
Here $B^m$ denotes the memory budget for modality $m$. 
This training-free update keeps long-horizon inference bounded while retaining a compact approximation to full-history attention.

%% file: section/experimets.tex
\section{Experiments}

\begin{table}[t]
  \caption{Comparison on NAVSIM v1. ${}^*$: results with imitation learning. $\dagger$: trained with multiple trajectory anchors from \citep{li2024hydra}. MV: multi-view cameras; SV: single-view camera; L: LiDAR.}
  \label{tab:navsim_v1}
  \centering
  \resizebox{\linewidth}{!}{
  \begin{tabular}{lcc|cccccc}
    \toprule
    Method & Ref & Sensors & NC $\uparrow$ & DAC $\uparrow$ & TTC $\uparrow$ & C. $\uparrow$ & EP $\uparrow$ & \cellcolor{gray!20}PDMS $\uparrow$ \\
    \midrule
    Human & -- & -- & 100 & 100 & 100 & 99.9 & 87.5 & \cellcolor{gray!20}94.8 \\
    \midrule
    UniAD~\cite{hu2023planning} & CVPR'23 & MV  & 97.8 & 91.9 & 92.9 & \textbf{100.0} & 78.8 & \cellcolor{gray!20}83.4 \\
    TransFuser~\cite{chitta2022transfuser} & TPAMI'23 & MV \& L & 97.7 & 92.8 & 92.8 & \textbf{100.0} & 79.2 & \cellcolor{gray!20}84.0 \\
    PARA-Drive~\cite{weng2024drive} & CVPR'24 & MV & 97.9 & 92.4 & 93.0 & 99.8 & 79.3 & \cellcolor{gray!20}84.0 \\
    LAW~\cite{li2025enhancing} & ICLR'25 & SV & 96.4 & 95.4 & 88.7 & 99.9 & 81.7 & \cellcolor{gray!20}84.6 \\
    DiffusionDrive~\cite{liao2025diffusiondrive} & CVPR'25 & MV \& L & 98.2 & 96.2 & 94.7 & \textbf{100.0} & 82.2 & \cellcolor{gray!20}88.1 \\
    WoTE~\cite{li2025end} & ICCV'25 & MV \& L & 98.5 & 96.8 & 94.4 & 99.9 & 81.9 & \cellcolor{gray!20}88.3 \\
    \midrule

    \multicolumn{9}{l}{\textit{\textbf{VLA-based Methods}}} \\
    ReCogDrive${}^*$~\cite{li2025recogdrive} & ICLR'26 & MV & 98.1 & 94.7 & 94.2 & 100.0 & 80.9 & \cellcolor{gray!20}86.5 \\
    DriveVLA-W0~\cite{li2026drivevlaw} & ICLR'26 & SV & \textbf{98.7} & 96.2 & 95.5 & \textbf{100.0} & 82.2 & \cellcolor{gray!20}88.4 \\
    AutoVLA~\cite{zhou2026autovla} & NeurIPS'25 & MV & 98.4 & 95.6 & \textbf{98.0} & 99.9 & 81.9 & \cellcolor{gray!20}89.1 \\
    DriveDreamer-Policy~\cite{zhou2026drivedreamer} & arXiv'26 & MV & 98.4 & 97.1 & 95.1 & \textbf{100.0} & 83.5 & \cellcolor{gray!20}89.2 \\
    \textcolor{gray!65}{DriveVLA-W0$\dagger$~\cite{li2026drivevlaw}} & \textcolor{gray!65}{ICLR'26} & \textcolor{gray!65}{SV} & \textcolor{gray!65}{98.7} & \textcolor{gray!65}{\textbf{99.1}} & \textcolor{gray!65}{95.3} & \textcolor{gray!65}{99.3} & \textcolor{gray!65}{83.3} & \cellcolor{gray!10}\textcolor{gray!65}{90.2} \\

    \midrule
    \multicolumn{9}{l}{\textit{\textbf{WA-based Methods}}} \\
    Epona~\cite{zhang2025epona} & ICCV'25 & SV & 97.9 & 95.1 & 93.8 & 99.9 & 80.4 & \cellcolor{gray!20}86.2 \\
    WorldDrive~\cite{gui2026bridging} & arXiv'26 & SV & 98.4 & 95.8 & 95.2 & 99.8 & 83.3 & \cellcolor{gray!20}89.0 \\
    DriveWAM & -- & SV & 98.3 & \textbf{98.1} & 95.2 & \textbf{100.0} & \textbf{84.3} & \cellcolor{gray!20}\textbf{90.1}\\
    \bottomrule
  \end{tabular}
  }
  \vspace{-10pt}
\end{table}

\begin{table}[t]
  \caption{Comparison on our curated 1,000-clip test subset of PhysicalAI-Autonomous-Vehicles benchmark. \# Params denotes the number of model parameters. SV: single-view camera. ${}^*$: evaluated using the released checkpoint, which only supports up to 3s prediction.
}
  \label{tab:physicalai}
  \centering
  \resizebox{0.95\linewidth}{!}{
  \begin{tabular}{lccc|>{\columncolor{gray!20}}c>{\columncolor{gray!20}}c>{\columncolor{gray!20}}c>{\columncolor{gray!20}}c}
    \toprule
    Method & Source & Sensors & \# Params & ADE@3s $\downarrow$ & FDE@3s $\downarrow$ & ADE@4s $\downarrow$ & FDE@4s $\downarrow$ \\
    \midrule
    VaVAM${}^*$~\cite{vavam2025} & Valeo & SV & 1.3B &  2.31 & 4.32 & -- & --\\
    Alpamayo-1.5~\cite{wang2025alpamayo} & NVIDIA & SV & 10B & 0.80 & 2.31 & 1.44 & 4.18 \\
    \midrule
    DriveWAM & -- & SV & 5B + 8B & \textbf{0.47} & \textbf{1.35} & \textbf{0.83} & \textbf{2.47} \\
    \bottomrule
  \end{tabular}
  }
  \vspace{-10pt}
\end{table}

\begin{table}[t]
  \caption{Ablation of scene-evolving (SE) driving guidance under different training data scales on the PhysicalAI-Autonomous-Vehicles benchmark. \xmark: fixed global prompt as text conditioning.}
  \label{tab:ablation_guidance_scaling}
  \centering
  \resizebox{0.57\linewidth}{!}{
  \begin{tabular}{lcc|cc}
    \toprule
    \# Clips & \# Iters & SE Guidance & ADE@4s $\downarrow$ & FDE@4s $\downarrow$ \\
    \midrule
    4k   & 50k & \xmark & 1.21 & 3.65 \\
    4k   & 50k & \cmark & \textbf{1.01} & \textbf{2.95} \\
    \midrule
    20k  & 50k & \xmark & 0.95 & 2.94 \\
    20k  & 50k & \cmark & \textbf{0.94} & \textbf{2.65} \\
    \midrule
    100k & 50k & \xmark & 0.92 & 2.75 \\
    100k & 50k & \cmark & \textbf{0.83} & \textbf{2.47} \\
    \bottomrule
  \end{tabular}
  }
\end{table}

\begin{table}[t]
\centering
\vspace{-10pt}
\begin{minipage}{0.45\linewidth}
  \centering
  \caption{Ablation of video backbone initialization and joint video supervision. All models are trained on 100k clips for 50k iterations.}
  \label{tab:ablation_pretrain_video}
  \resizebox{\linewidth}{!}{
  \begin{tabular}{cc|cc}
    \toprule
    Pretrained init. & Video sup. & ADE@4s $\downarrow$ & FDE@4s $\downarrow$ \\
    \midrule
    \xmark & \cmark & 1.10 & 3.26 \\
    \cmark & \xmark & 1.23 & 3.79 \\
    \cmark & \cmark & \textbf{0.83} & \textbf{2.47} \\
    \bottomrule
  \end{tabular}
  }
\end{minipage}
\hfill
\begin{minipage}{0.52\linewidth}
  \centering
  \caption{Ablation of KV memory strategies. ADE/FDE are measured on 20s clips, while KV memory and GFLOPs are profiled under a 300s clip.}
  \label{tab:ablation_kv_memory}
  \resizebox{\linewidth}{!}{
  \begin{tabular}{l|cc|cc}
    \toprule
    KV memory & ADE@4s $\downarrow$ & FDE@4s $\downarrow$ & Mem. (GB) $\downarrow$ & GFLOPs $\downarrow$ \\
    \midrule
    Full & \textbf{0.83} & \textbf{2.47} & 3.07 & 17.37 \\
    FIFO & 1.40 & 3.47 & \textbf{0.25} & \textbf{1.05} \\
    Selective & 0.89 & 2.52 & \textbf{0.25} & 1.44 \\
    \bottomrule
  \end{tabular}
  }
\end{minipage}
\vspace{-10pt}
\end{table}

\subsection{Datasets}

\textbf{NAVSIM}~\citep{dauner2024navsim} is a standard end-to-end planning benchmark derived from OpenScene~\cite{openscene2023,karnchanachari2024towards}, with 103k trainval samples and 12k test samples. Following the standard NAVSIM protocol, we report No at-fault Collisions (NC), Drivable Area Compliance (DAC), Time-To-Collision (TTC), Comfort (C.), Ego Progress (EP), and the overall Predictive Driver Model Score (PDMS).

\textbf{PhysicalAI-Autonomous-Vehicles} is a large-scale real-world driving benchmark released with Alpamayo-R1~\citep{wang2025alpamayo}. It contains approximately 1,700 hours of driving logs, organized into 306,152 clips of 20 seconds each, with 153,625 clips for training, 90,928 for validation, and 61,599 for testing. We use the front-view camera stream and ego-motion labels. To focus on non-trivial driving scenarios, we use a VLM to tag each clip with a scene description and filter out simple scenes. Finally, we select 100k clips from the training split, and construct a curated 1,000-clip test subset from the test split. Details of the filtering procedure are provided in Appendix~\ref{app:dataset}. We report Average Displacement Error (ADE) and Final Displacement Error (FDE) over 3-second and 4-second future trajectories.

\subsection{Implementation Details}
We build DriveWAM based on the code framework of~\citep{li2026causal}. DriveWAM uses Wan2.2-TI2V-5B~\cite{wan2025wan} as the video backbone, initialized from the base checkpoint released by~\citep{li2026causal}. Unless otherwise specified, we fine-tune the full video diffusion transformer together with the newly introduced action and ego-state modules. The action encoder $E_a$ and action decoder $D_a$ are implemented as MLPs with hidden dimension 3072, and the ego-state features are encoded by a separate MLP. The scene-evolving guidance is generated by a frozen Qwen3-VL-8B~\citep{Qwen3-VL}, which is queried once per chunk. Details of the VLM prompt template are provided in Appendix~\ref{app:vlm_guidance}.

All models are trained at $256{\times}448$ resolution on 48 NVIDIA H20 GPUs. We use AdamW~\citep{loshchilov2018decoupled} with $\beta=(0.9,0.95)$, weight decay 0.1, learning rate $1{\times}10^{-5}$, and per-device batch size 1. The action loss weight is set to $\beta_a=1.0$. DriveWAM uses a 4-second chunk for video-action generation. On NAVSIM, we train for 100k iterations and decay the learning rate by a factor of 0.5 at 50k, 70k, and 90k iterations. Each sample uses the current frame as the condition and predicts a 4-second future horizon at 1\,Hz. Since NAVSIM provides a single future planning horizon per sample, this setting reduces to one chunk-level prediction. On the PhysicalAI-Autonomous-Vehicles benchmark, we train for 50k iterations. Each training sample is a 12-second segment randomly cropped from a 20-second clip. The video stream is downsampled to 1\,Hz, while ego actions remain at 10\,Hz.

For inference, following~\citep{li2026causal}, we use an Euler ODE solver with $3$ steps for video tokens and $10$ steps for action tokens. The video solver integrates the flow trajectory from $\tau=1$ to $\tau=0.6$, while the action solver integrates from $\tau=1$ to $\tau=0$. For selective KV memory, we follow FlowCache~\citep{ma2026flow} and set $\lambda=0.07$. The video and action cache capacities are set to 448 and 160 tokens, respectively.

\subsection{Main Results}
\textbf{NAVSIM.} We compare DriveWAM against state-of-the-art end-to-end planners on NAVSIM v1, including classical end-to-end pipelines~\citep{hu2023planning,chitta2022transfuser,weng2024drive,li2025enhancing,liao2025diffusiondrive,li2025end}, VLA-based policies~\citep{li2025recogdrive,li2026drivevlaw,zhou2026autovla,zhou2026drivedreamer}, and WA-based methods~\citep{zhang2025epona,gui2026bridging}. As shown in Table~\ref{tab:navsim_v1}, DriveWAM achieves a PDMS of $90.1$ using only a single front-view camera, outperforming all competing methods under comparable training settings. We attribute this to the underlying video generative backbone, which provides effective spatio-temporal priors for modeling scene geometry, motion dynamics, and fine-grained action prediction.

\textbf{PhysicalAI-Autonomous-Vehicles.} We evaluate DriveWAM on the large-scale PhysicalAI-Autonomous-Vehicles benchmark, comparing against the WA-based VaVAM~\citep{vavam2025}, trained on approximately 1{,}700 hours of OpenDV~\cite{yang2024generalized} driving data, and the VLA-based Alpamayo-1.5~\citep{wang2025alpamayo}, trained on roughly 80{,}000 hours of data containing the PhysicalAI-Autonomous-Vehicles training set. For consistency, all methods use only the front-view camera input and output a single trajectory at inference. As reported in Table~\ref{tab:physicalai}, DriveWAM achieves ADE/FDE of $0.47$/$1.35$ at 3 seconds and $0.83$/$2.47$ at 4 seconds, substantially outperforming both baselines.

\begin{figure}
  \centering
  \includegraphics[width=0.99\linewidth]{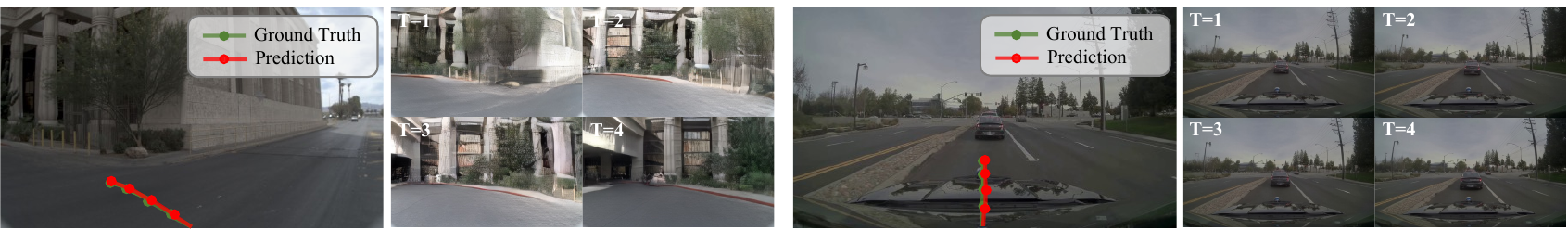}
\caption{Qualitative results on NAVSIM (left) and PhysicalAI-Autonomous-Vehicles benchmark (right). The predicted ego trajectories are consistent with the jointly generated future scenes.}
  \label{fig:vis}
  \vspace{-10pt}
\end{figure}

\textbf{Qualitative results.} Figure~\ref{fig:vis} visualizes future scenes and ego trajectories jointly generated by \methodname. Additional qualitative examples are provided in Appendix~\ref{app:vis}.

\subsection{Ablation Study}
We conduct ablation studies on the PhysicalAI-Autonomous-Vehicles benchmark to investigate the individual components of DriveWAM. Unless otherwise noted, all ablation models are trained on 100k clips for 50k iterations under the same optimization settings as in the main results.

\textbf{Scene-evolving Driving Guidance.} Table~\ref{tab:ablation_guidance_scaling} studies the contribution of injecting chunk-specific VLM guidance. Replacing the global prompt with scene-evolving guidance consistently improves trajectory prediction at every training data scale, reducing ADE@4s from $1.21$ to $1.01$ with 4k clips and from $0.92$ to $0.83$ with 100k clips, while also yielding consistent reductions in FDE@4s. These results indicate that high-level scene reasoning provides a complementary semantic conditioning to the low-level WA backbone. We also observe that the benefit does not vanish as training data grows. Appendix~\ref{app:vlm_guidance} provides qualitative examples of guidance evolving with scene context and route intent.

\begin{wrapfigure}{r}{0.4\linewidth}
  \vspace{-1em}
  \centering
  \includegraphics[width=\linewidth]{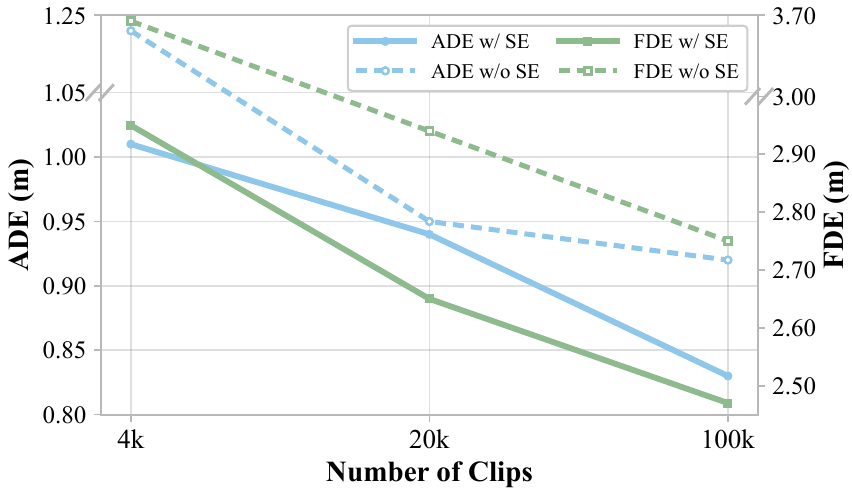}
  \vspace{-1.5em}
  \caption{Data scaling on PhysicalAI-Autonomous-Vehicles.}
  \label{fig:data_scaling}
  \vspace{-1em}
\end{wrapfigure}

\textbf{Data Scaling.} We investigate the data scalability of DriveWAM by varying the training set size from 4k to 20k and 100k clips under a fixed 50k-iteration training procedure. As shown in Table~\ref{tab:ablation_guidance_scaling} and Figure~\ref{fig:data_scaling}, both ADE@4s and FDE@4s improve significantly with more data, regardless of whether scene-evolving guidance is applied. This consistent scaling trend reflects the effectiveness of the video-action modeling as a scalable policy foundation, and suggests that DriveWAM has not yet saturated at the current data scale.

\textbf{Video Foundation Model Adaptation.} We ablate DriveWAM's capability by removing the pretrained video-backbone initialization and the joint video flow-matching supervision. As reported in Table~\ref{tab:ablation_pretrain_video}, training entirely from scratch removes the large-scale spatio-temporal priors inherited from video pretraining, and degrades ADE@4s/FDE@4s to $1.10$/$3.26$. Initializing from the pretrained backbone but removing video supervision also performs poorly, yielding $1.23$/$3.79$, suggesting that action-only adaptation fails to preserve the generative video priors needed for WA policy learning. The full configuration combines pretrained initialization with joint video-action flow-matching supervision and achieves the best performance.

\textbf{Selective KV Memory.} Table~\ref{tab:ablation_kv_memory} compares three inference-time memory strategies for autoregressive rollout. Full KV caching retains the entire video-action history, while FIFO and our selective KV memory operate under the fixed-size cache budget. As shown in $1^{st}$ and $3^{rd}$ rows, selective KV memory largely closes the accuracy gap to full caching, achieving $0.89$/$2.52$ ADE@4s/FDE@4s, while FIFO degrades substantially to $1.40$/$3.47$. To examine the long-horizon overhead of the memory module, we further profile each strategy on a 300-second rollout, reporting KV memory summed over all DiT layers and attention GFLOPs of one causal self-attention layer. As presented in Table~\ref{tab:ablation_kv_memory}, full caching requires $3.07$ GB memory and $17.37$ GFLOPs per step, whereas selective KV memory reduces them to $0.25$ GB and $1.44$ GFLOPs, yielding over $12{\times}$ reductions.

\section{Conclusion}
We present \methodname, a unified world-action policy that adapts a pretrained video foundation model directly into an end-to-end driving policy. \methodname introduces scene-evolving driving guidance that injects chunk-specific semantic intent through temporally localized cross-attention, and selective KV memory that maintains modality-aware video and action memory pools via relevance-redundancy selection at inference time. Experiments on NAVSIM and the PhysicalAI-Autonomous-Vehicles benchmark show that \methodname achieves strong planning performance, and a data-scaling study from 4k to 100k clips further confirms its scalability.

%% file: section/appendix.tex
\section{Dataset Curation}
\label{app:dataset}
The PhysicalAI-Autonomous-Vehicles benchmark contains roughly 1{,}700 hours of driving organized into 306{,}152 20-second clips. To focus evaluation and training on non-trivial driving scenarios, we tag every clip with a frozen Qwen3-VL-8B~\citep{Qwen3-VL} and use the resulting tags to construct a 100k-clip training subset, and a curated 1{,}000-clip test subset with balanced coverage of rare and ordinary scenarios.

\begin{figure}
  \centering
  \includegraphics[width=0.999\linewidth]{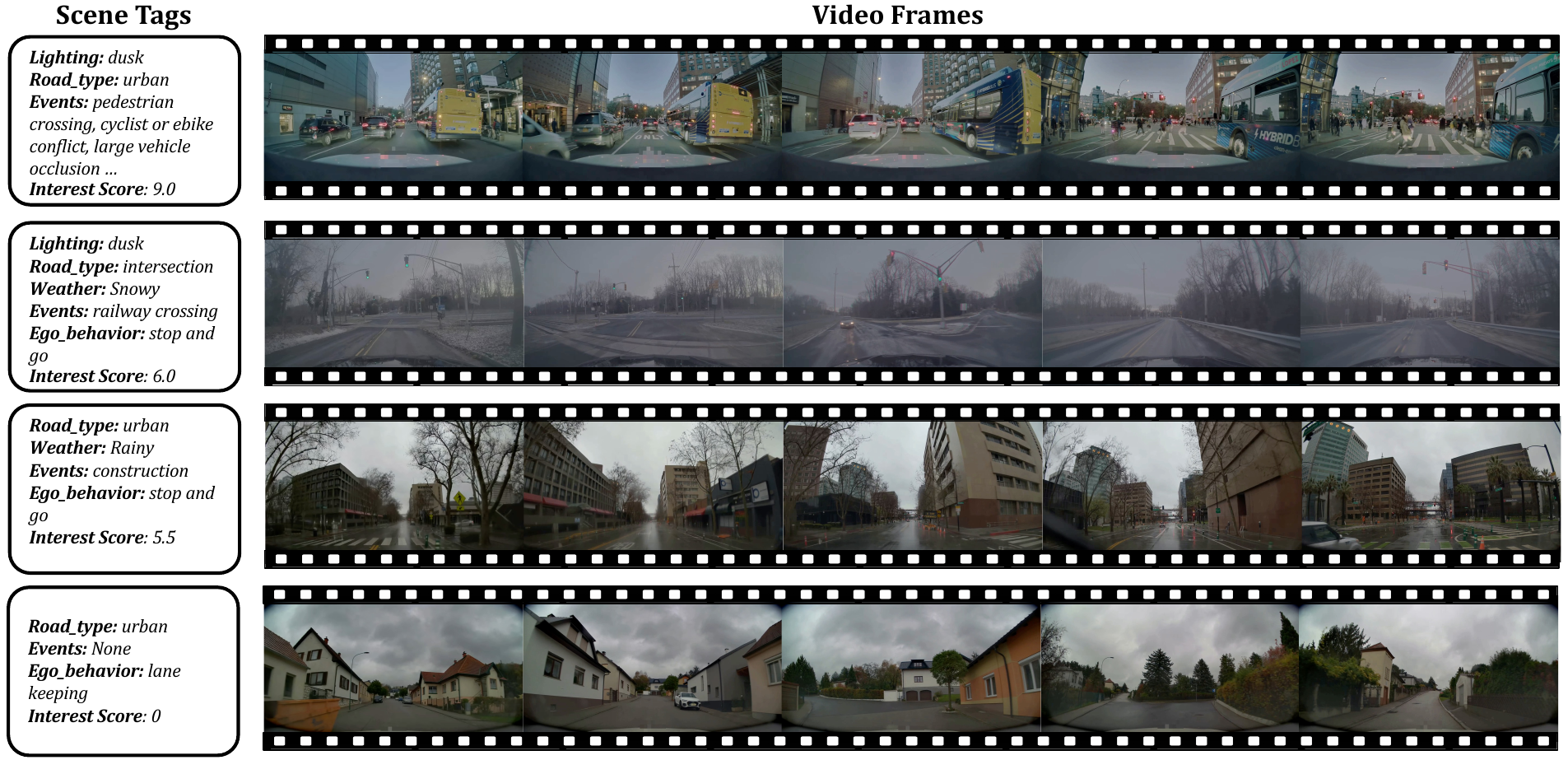}
  \caption{Representative scene tagging results for dataset curation. For each clip, the left panel shows Qwen3-VL-8B detected scene attributes, events, and the resulting interest score, while the right panel shows sampled front-view frames. High-score clips capture rare or interaction-rich scenarios, whereas low-score clips represent ordinary driving.}
  \label{fig:appendix_tag}
\end{figure}

\textbf{Scene tagging.} For each clip, we uniformly sample 20 frames from the front-view stream and pass them to Qwen3-VL-8B with four structured prompts. Each prompt focuses on one facet of driving complexity:
\begin{itemize}
    \item \textbf{Scene attributes}: weather (clear/rainy/snowy/foggy), lighting (day/dusk/night/tunnel transition/strong backlight), road type (urban/highway/ramp/intersection/etc.), traffic density, and ego behavior.
    \item \textbf{Vulnerable road-user events}: whether the scene contains pedestrian crossing, jaywalking, occluded pedestrian popout, child or elderly participants, cyclist conflict, or crowd.
    \item \textbf{Vehicle interaction events}: whether the scene contains cut-in, cut-out, sudden braking ahead, wrong-way vehicle, large-vehicle occlusion, emergency vehicle, door opening, or stopped/broken vehicle.
    \item \textbf{Intersection and long-tail events}: whether the scene contains unprotected left turn, roundabout, irregular intersection, traffic-police gesture, road debris, accident scene, construction, animal on road, water puddle, or railway crossing, together with the traffic-light state.
\end{itemize}
The four prompts are run sequentially on the same sampled frames and merged into a single per-clip record. We additionally compute a scalar interest score by summing rule-based weights over detected event tags. In practice, rare or safety-critical events receive larger weights, e.g., accident scenes ($5.0$), occluded pedestrian popouts ($4.0$), animals on the road ($3.5$), and traffic-police gestures ($3.0$), while frequent or lower-impact attributes receive smaller weights ($0.5$--$1.5$). Figure \ref{fig:appendix_tag} shows representative tagged clips with their detected attributes and interest scores.

\textbf{Training subset.} The training subset is curated from the tagged training split through a two-stage procedure. We first retain all clips with interest score no smaller than $2.0$, preserving rare-event and interaction-rich cases. We then uniformly sample 50\% of the remaining lower-score clips, so that ordinary driving scenarios are still represented without dominating the training distribution. For the data-scaling study, we sample 20k and 4k subsets from this 100k subset.

\textbf{Test subset.} The test subset contains 1,000 clips and is constructed from the tagged test split to cover both long-tail and ordinary driving scenarios. We combine three sources: 
\begin{itemize}
  \item \textbf{Rare-event clips}: a tag is treated as rare if it appears in fewer than 1\% of the test clips. For each rare tag, we select up to 30 top-scoring clips that contain it, covering events such as accident scenes, animals on the road, occluded pedestrian popouts, traffic-police gestures, and railway crossings.
  \item \textbf{High-interest clips}: clips above the 75th percentile of the interest-score distribution are grouped by weather, lighting, and road type. We assign an approximately equal quota to each group and select the highest-scoring clips within each group until the target size is reached.
  \item \textbf{Common-scene clips}: 200 clips uniformly sampled from below the high-interest threshold to serve as ordinary-driving controls.
\end{itemize}
The selected clips are merged to form the final 1,000-clip test set.

\begin{figure}[t]
    \centering
    \includegraphics[width=\linewidth]{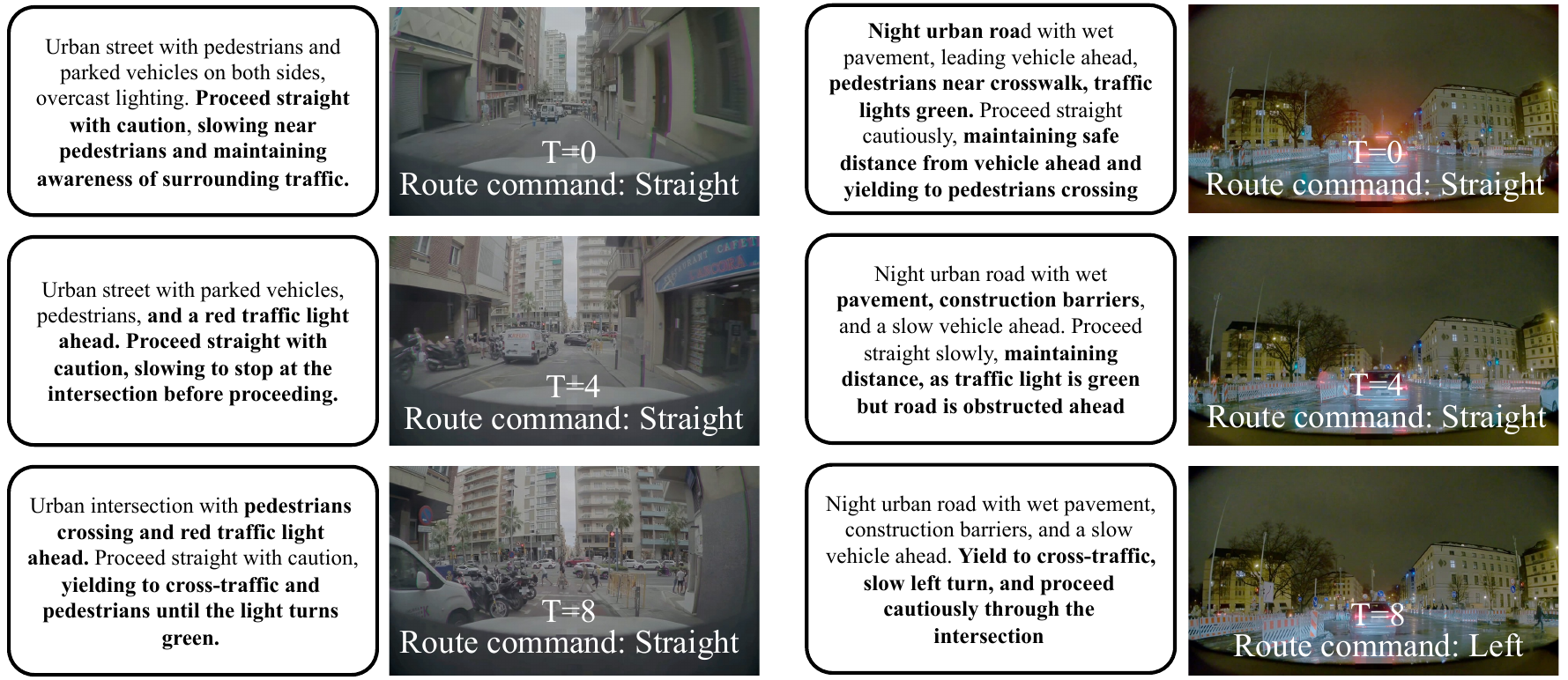}
    \caption{
    Examples of scene-evolving VLM guidance. The guidance adapts to changing scene context and route intent, such as pedestrians, traffic lights, construction barriers.}
    \label{fig:vlm_guidance_examples}
\end{figure}

\section{VLM Guidance Details}
\label{app:vlm_guidance}
This section details the pipeline that produces the chunk-specific guidance $g_k$ used in Sec.~\ref{sec:vlm_guidance}. The pipeline operates in two stages. First, we classify the route of each upcoming 4-second chunk from ground-truth ego pose, producing a route command. Second, we prompt a frozen Qwen3-VL-8B with the route command, the front-camera frame at the end of the latest chunk, and a BEV visualization of the ego trajectory from the previous 4-second chunk, asking it to produce a concise two-sentence guidance for the upcoming chunk. Figure~\ref{fig:vlm_guidance_examples} shows representative guidance examples, where the generated text evolves with the latest observation and route command.

\paragraph{Route command.}
Each chunk is assigned a high-level route command from \{\texttt{straight}, \texttt{left}, \texttt{right}\}. As explicit route annotations are unavailable, we construct this coarse command from the route/ego-yaw change for labeling purposes. Specifically, this command is derived from the yaw change of the ego vehicle over the chunk. Let $R_0$ and $R_1$ denote the ego rotations at the beginning and end of a chunk. We compute the relative yaw from $R_0^\top R_1$ and assign the command as \texttt{left} if the yaw change is larger than $15^\circ$, \texttt{right} if it is smaller than $-15^\circ$, and \texttt{straight} otherwise. The command only specifies directional intent and does not contain future positions, velocities, distances, or trajectory coordinates.

\paragraph{Prompt template.}
The prompt template used for chunk-level guidance generation is shown below. The route command and visual inputs are filled at runtime.

\begin{tcolorbox}[promptbox,title={Prompt template for VLM guidance}]
\small
You are a navigation assistant for an autonomous-driving dataset. You generate short navigation guidance for an upcoming 4-second driving window. You do not see the future window; you only see the current road conditions, the recent ego trajectory when available, and a high-level route command.

\medskip
\textbf{Inputs.}
\begin{itemize}
\setlength{\itemsep}{1pt}
\setlength{\parsep}{0pt}
\item Route command for the upcoming window: \texttt{straight}, \texttt{left}, or \texttt{right}. Treat it as an authoritative navigation instruction.
\item Latest causally available front-camera frame before the upcoming 4-second window, showing the current road conditions.
\item BEV trajectory map of the previous 4-second window when available, showing recent ego motion and speed.
\end{itemize}

\medskip
\textbf{Output format.}
Output exactly two sentences, under 50 words in total. Do not use labels, bullets, markdown, or extra paragraphs. Use present tense only.

\medskip
\textbf{Sentence 1.}
Describe the current road context visible in the provided frame, including road type, traffic participants, traffic-light state if visible, weather, and lighting.

\medskip
\textbf{Sentence 2.}
Describe the ego navigation guidance for the upcoming window. Jointly reason from the road context, the previous BEV trajectory, and the route command. The direction must be consistent with the route command, and the caution level should reflect the traffic conditions. Capture required interactions such as yielding, waiting, or merging. Use qualitative language only; do not include numbers, units, distances, coordinates, or low-level trajectory values.
\end{tcolorbox}

\section{Efficiency Analysis}

We analyze the per-chunk inference cost of DriveWAM and compare it against Alpamayo-1.5~\citep{wang2025alpamayo} on a single NVIDIA H20 GPU. As shown in Table~\ref{tab:efficiency}, each inference pass consists of three stages: (1) VLM guidance generation, (2) video generation, and (3) action denoising.

\textbf{VLM guidance.}
DriveWAM queries a frozen Qwen3-VL-8B once per 4-second chunk, taking $125$\,ms with the default vLLM compilation. Because the guidance is generated at the chunk boundary rather than per frame, the cost is amortized over the entire chunk. Alpamayo-1.5 processes a substantially larger number of visual tokens per query, which accounts for its higher VLM latency of $570$\,ms.

\textbf{Video generation.}
DriveWAM generates a 4-second video clip using a 3-step Euler ODE solver over the video tokens, taking $372$\,ms. Alpamayo-1.5 does not perform explicit video generation.

\textbf{Action denoising.}
By default, DriveWAM uses 10 denoising steps for action tokens, taking $765$\,ms. We find that reducing the steps from 10 to 5 incurs negligible change in trajectory metrics, while reducing action denoising time to $374$\,ms. The 5-step variant (DriveWAM$^*$) brings the total per-chunk cost to approximately $871$\,ms, comparable to Alpamayo-1.5's $900$\,ms, while additionally producing a jointly generated future video.

\begin{table}[h]
\centering
\small
\setlength{\tabcolsep}{8pt}
\begin{tabular}{lccccc}
\toprule
Method & VLM (ms) & Video Gen (ms) & Action (ms) & ADE@4s $\downarrow$ & FDE@4s $\downarrow$ \\
\midrule
Alpamayo-1.5     & 570 & ---  & 330  & 1.44 & 4.18 \\
DriveWAM (Ours)  & 125 & 372  & 765  & \textbf{0.83} & 2.47 \\
DriveWAM$^*$ (Ours)  & 125 & 372  & 374 & 0.84 & \textbf{2.45} \\
\bottomrule
\end{tabular}
\caption{Per-chunk inference cost and trajectory prediction accuracy on a single H20 GPU. $^*$ indicates action denoising steps reduced from 10 to 5.}
\label{tab:efficiency}
\end{table}

\section{Additional Qualitative Results}
\label{app:vis}

\begin{figure}[htbp]
  \centering
  \includegraphics[width=0.999\linewidth]{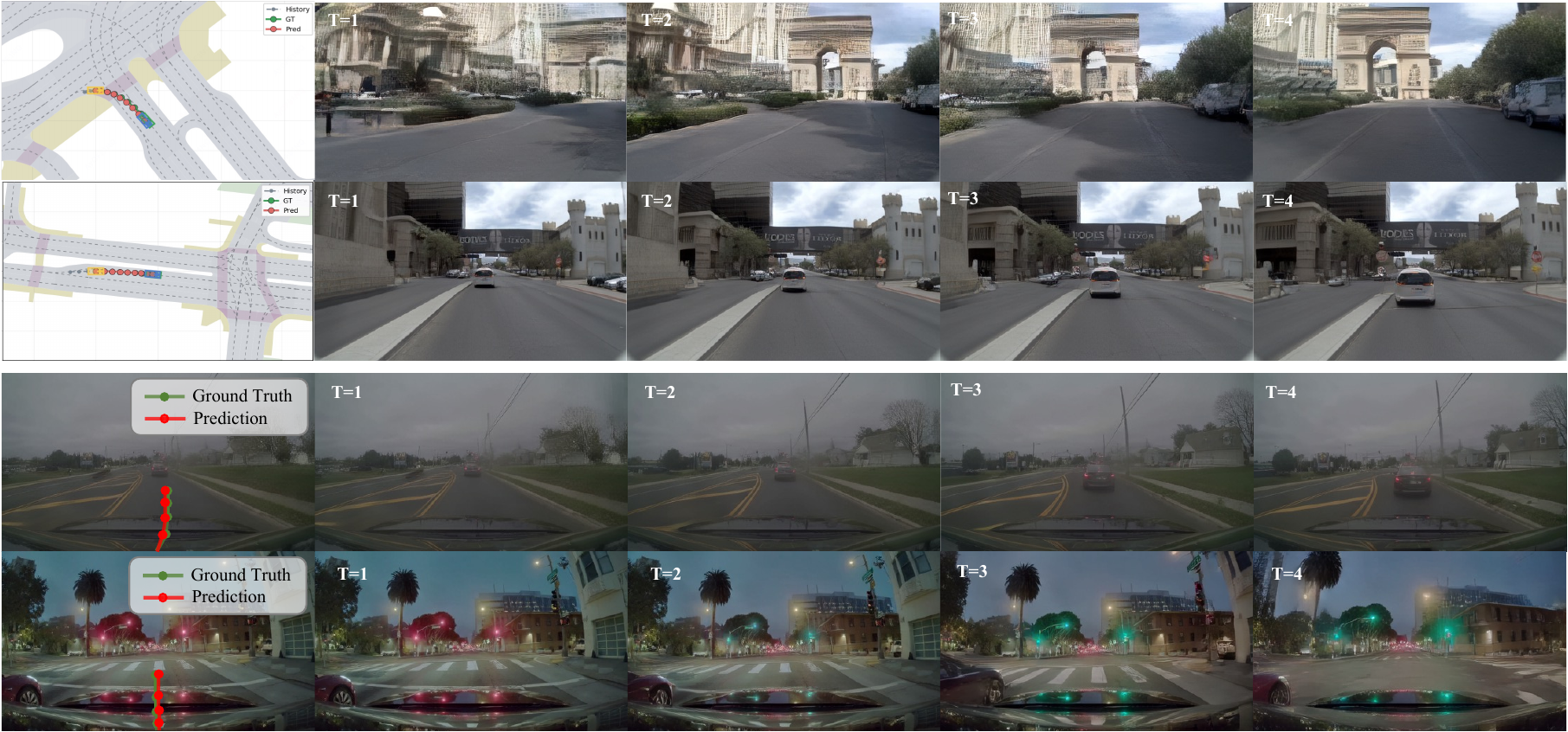}
  \caption{Qualitative results on NAVSIM (top two rows) and PhysicalAI-Autonomous-Vehicles (bottom two rows) benchmarks. Each row shows the predicted ego trajectory alongside the jointly generated future frames at T=1,2,3,4. }
  \label{fig:more_results}
\end{figure}

We present additional qualitative results to complement the main-paper visualization. Figure~\ref{fig:more_results} shows representative examples from both NAVSIM and the PhysicalAI-Autonomous-Vehicles benchmark, spanning driving conditions and road layouts.

\textbf{NAVSIM qualitative results.}
Each example shows a BEV map on the left, where the red trajectory is the DriveWAM prediction and the blue trajectory is the ground-truth. The yellow vehicle icon denotes the starting ego, the blue vehicle icon denotes the predicted ending ego, and the green vehicle icon denotes the ground-truth ending ego. In both cases, the predicted trajectory aligns closely with the ground truth despite the complexity of the surroundings, and the generated video maintains photometric and geometric consistency across the four future timesteps.

\textbf{PhysicalAI-Autonomous-Vehicles qualitative results.}
Each example overlays the ground-truth and predicted ego trajectories on the current front-view frame.
These results are consistent with the strong quantitative performance and further demonstrate that the joint video-action generation provides a coherent, physically plausible world model that supports accurate long-horizon planning across diverse real-world conditions.